\documentclass{Interspeech2024}

\usepackage{graphicx}
\usepackage{subcaption}

\interspeechcameraready

\title{A Functional Trade-off between Prosodic and Semantic Cues in Conveying Sarcasm}

\name[affiliation={1}]{Zhu}{Li}
\name[affiliation={1}]{Xiyuan}{Gao}
\name[affiliation={2}]{Yuqing}{Zhang}
\name[affiliation={1}]{Shekhar}{Nayak}
\name[affiliation={1}]{Matt}{Coler}

\address{
  $^1$Campus Frysl\^{a}n, University of Groningen, the Netherlands \\
  $^2$Center for Language and Cognition, University of Groningen, the Netherlands
  }
\email{\{zhu.li, xiyuan.gao, yuqing.zhang, s.nayak, m.coler\}@rug.nl}

\keywords{sarcasm, functional trade-off, prosodic cues, 
semantic cues}

\begin{document}

\maketitle
\begin{abstract} 
    This study investigates the acoustic features of sarcasm and disentangles the interplay between the propensity of an utterance being used sarcastically and the presence of prosodic cues signaling sarcasm. Using a dataset of sarcastic utterances compiled from television shows, we analyze the prosodic features within utterances and key phrases belonging to three distinct sarcasm categories (embedded, propositional, and illocutionary), which vary in the degree of semantic cues present, and compare them to neutral expressions. Results show that in phrases where the sarcastic meaning is salient from the semantics, the prosodic cues are less relevant than when the sarcastic meaning is not evident from the semantics, suggesting a trade-off between prosodic and semantic cues of sarcasm at the phrase level. These findings highlight a lessened reliance on prosodic modulation in semantically dense sarcastic expressions and a nuanced interaction that shapes the communication of sarcastic intent. 

\end{abstract}

\section{Introduction}
A sarcastic sentence is one in which a speaker says the opposite of what they mean, adding irony and humor to language \cite{sperber1981irony, grice1975logic}. Although it is used frequently, sarcasm poses distinct challenges in distinguishing it from neutral speech. Understanding the ways sarcasm is expressed and meant is crucial, not just for theory but for use in interactive spoken dialogue systems.

Our investigation into previous research reveals diverse outcomes influenced by the languages under study and the methods employed to elicit sarcasm. In languages like English, French, Italian, German, and Cantonese, changes in the basic pitch, or fundamental frequency ($F0$) are linked with sarcasm \cite{scharrer2011voice, rockwell2000lower, cheang2008sound, anolli2000irony, cheang2009acoustic, loevenbruck2013prosodic, chen2018s}. However, its usage varies across languages. Studies in Cantonese, Italian, and French report an increase in mean $F0$ to mark sarcasm, while studies in German demonstrated a reduced $F0$ range \cite{scharrer2011voice, cheang2009acoustic, anolli2000irony, loevenbruck2013prosodic}. The divergence is particularly notable in studies focusing on English. \cite{cheang2009acoustic} reported decreased mean $F0$, $F0$ range, and $F0$ standard deviation as markers of sarcasm, while \cite{rockwell2000lower} observed an increased mean $F0$ and $F0$ range as cues of sarcastic speech. The mixed results are further complicated by different methods used to elicit sarcasm. For instance, distinct from the aforementioned studies, \cite{bryant2005there} observed that spontaneous ironic speech, extracted from talk radio shows, offered minimal evidence for distinguishable features. Recognizing the methodological diversity in previous sarcasm research, we integrate these varied approaches with a comprehensive methodology that combines elements from spontaneous and elicited sarcasm studies. This mitigates the disparities observed in prior findings, offering a holistic understanding of acoustic sarcasm markers.  Apart from $F0$, reductions in speech rate and the lengthening of syllables or entire utterances emerge as consistent features of sarcasm across languages. Changes in intensity have also been identified in sarcastic utterances across languages \cite{scharrer2011voice, anolli2002blame}. However, diversity is noticeable even within English; \cite{rockwell2000lower} reported increased intensity in sarcastic utterances, while \cite{cheang2008sound} found no corresponding increase compared to literal utterances.

Another confounding factor is the different types of sarcasm. For example, \cite{bryant2005there} differentiated between ``dry sarcasm", where sarcasm is inferred only from context, and ``dripping sarcasm", marked by explicit word choice and unambiguous prosody, finding that ``dry sarcasm" featured a lower mean syllabic duration while dripping sarcasm exhibited a higher one. \cite{cheang2008sound} further compared acoustic features in short phrases (i.e., ``Yeah, right.") and long sentences, uncovering that speech rate varies with sentence length to indicate sarcasm, with shorter sentences marked by a slower speech rate. Such studies underscore the interplay between the kind of sarcasm and the diversity of cues employed.


We unravel the complexities of sarcastic meaning by investigating whether modulations of $F0$, duration, and amplitude – parameters consistently associated with sarcastic speech across various languages – are intertwined with the level of context ambiguity behind sarcasm. We explore three major subtypes of sarcasm: embedded, propositional, and illocutionary. Each subtype is characterized by a distinct use of semantic cues, which are key elements our experiments aim to investigate for a deeper understanding of how sarcasm is conveyed. 
To ensure alignment with previous work, we focus on acoustic features that have shown significant correlation with sarcasm, including mean $F0$, $F0$ range, mean amplitude, and speaking rate.

The presence of semantic cues, often characterized by a mismatch between the context and the literal content, is a primary indicator of sarcasm.
To further explore the interaction between these two types of cues, we propose to investigate how they compensate for each other in conveying sarcasm. Given the varied reliance on prosodic and semantic cues across different instances of sarcasm, 
it is crucial to understand how these cues interact, particularly in contexts where one type of cue may be less apparent. This leads us to consider a potential reciprocal dynamic, where the absence or presence of one cue type influences the reliance on and manifestation of the other. Based on the assumption, we propose the following hypothesis (H1): 

H1: There is a reciprocal relationship between the use of prosodic and semantic cues in conveying sarcasm. In the absence of semantic cues, speakers emphasize prosodic cues to signal sarcasm. Conversely, when semantic indicators of sarcasm are evident, the reliance on prosodic cues diminishes.

While the role of semantic incongruity in sarcasm has been discussed, less attention has been paid to how specific prosodic markers are strategically used on key phrases within utterances to convey sarcasm. \cite{bryant2005there} proposed that the comprehension of sarcasm relies on an integration of linguistic content and local prosodic features, which manifest on a segmental level, contrasting with the global features that influence entire utterances. \cite{fowler1987talkers} elucidated how speakers accentuate the most informative words within an utterance by allocating greater articulatory effort to them. Based on these insights, our study suggests that speakers strategically target semantically critical phrases within an utterance to convey sarcasm effectively. We focus on the semantically critical phrases within utterances, 
and examine whether sarcasm is distinguishable from neutrality based solely on these key phrases. 
For instance, in the utterance ``It's just a privilege to watch your mind at work," the word \textit{privilege} holds the key to the sarcastic interpretation of the statement. The example provided illustrates this point but raises questions about the systematicity of this phenomenon across different instances of sarcasm. Therefore, 
we endeavor to dissect these occurrences, examining whether such prosodic emphasis on key phrases is a consistent feature across different sarcastic contexts. We propose hypothesis (H2):

H2: Prosodic features indicating sarcasm, such as changes in pitch or loudness, are predominantly found in specific phrases within a sentence that are crucial for imparting the sarcastic meaning. These key phrases serve as focal points for the application of prosodic sarcasm markers.

This paper delves into the intricate interplay between prosodic features and diverse types of sarcasm in spontaneous speech. Our contributions are:
\begin{itemize}
    \item We elucidate the nuances of sarcastic cues by revealing the interplay between prosodic cues of sarcasm and the various subtypes of sarcasm characterized by differing levels of context ambiguity.
    \item We underscore the importance of incorporating key phrases that convey sarcastic meaning into studies examining sarcasm in speech, enriching our understanding of how sarcasm is expressed 
    and perceived and presenting effective methods for examining intricate dialogue data.
    \item We provide empirical evidence that certain speech features serve as indicators of sarcasm in spontaneous speech.
    \item  We delineate the interplay between prosodic and semantic cues in detecting sarcasm, providing insights that enhance the accuracy of spoken dialogue systems.
\end{itemize} 

\section{Methodology}
\label{section:methodology}

\subsection{Data}

Our experiments leveraged the MUStARD++ dataset \cite{ray-etal-2022-multimodal}, comprising 1202 audiovisual utterances extracted from American TV sitcoms. These utterances are evenly divided between sarcastic and non-sarcastic instances and are accompanied by preceding conversations and speaker information. Each utterance is annotated with emotions, including valence and arousal, to capture the polarity and intensity of emotion. The sarcastic utterances are further classified into three distinct types of sarcasm:

\begin{itemize}
    \item Propositional Sarcasm, comprising 333 utterances, relies on contextual cues for sarcasm identification \cite{zvolenszky2012gricean}. For instance, without context, utterances such as ``It's my privilege" may not be readily recognized as sarcastic.
	\item Embedded Sarcasm, encompassing 87 utterances, contains inherent incongruities within the utterance itself, rendering the sarcasm evident through the text alone. For example, ``You could charge people money to punch you." 
	\item Illocutionary Sarcasm, represented by 178 utterances, utilizes non-verbal cues, such as prosodic features or visual cues, to convey sarcasm. In these instances, the spoken words may appear sincere, but the sarcasm is evident through non-textual signals \cite{zvolenszky2012gricean}. An example would be saying ``That's right" while rolling one's eyes or emphasizing the word ``right", where literal neutrality contrasts with non-textual sarcasm.
\end{itemize}

We applied voice fix tools \footnote{https://github.com/haoheliu/voicefixer} to remove background noise and laughter from the original audios. 

\subsection{Annotation}
 

Since sarcastic utterances in the dataset differed lexically from non-sarcastic ones at the utterance level, it raises questions about the reliability of the reported findings when directly comparing sarcastic and non-sarcastic utterances. Therefore, we aim to examine the prosodic expression of sarcasm not only at the utterance level but also in the key phrases within an utterance. To facilitate comparison between the acoustic features of key phrases in both sarcastic and neutral utterances, we assigned two graduate students specializing in linguistics to conduct the annotation process, which involved two main steps.

Firstly, we selected all sarcastic utterances from the Big Bang Theory (BBT) and instructed annotators to identify lexically meaningful words (mostly nouns, verbs, adjectives, adverbs, and other content words) and phrases (hereafter referred to as key phrases) within each sentence. 
Exclamations (e.g., ``oh", ``yeah") were also included, as they play an important role in conveying sarcasm.
Building upon the annotated key phrases, we conducted an extensive search of utterances in BBT transcripts containing the key phrases spoken by the same speaker.

Secondly, annotators labeled the curated utterances as neutral or non-neutral, guided by the videos of the utterance and its relevant context. 
Each annotator labeled the full set of utterances independently. Inter-annotation yielded substantial agreement (Kappa score: 0.6042). We only kept utterances labeled as neutral by both annotators. We exclude illocutionary sarcasm data (44 in total) signaled exclusively by video modality and only keep those signaled by audio modality.

\begin{table}[h!]
  \caption{Summary of Data}
  \label{tab:word_styles}
   \centering
  \begin{tabular}{lllll}
    \toprule
    \textbf{}   & \multicolumn{3}{c}{\textbf{Sarcastic}}      & \textbf{Non-Sarcastic} \\
     & \multicolumn{1}{c}{EMB} & \multicolumn{1}{c}{PRO} & \multicolumn{1}{c}{ILL}   & \\
    \midrule
    Utterance       & \multicolumn{1}{c}{$87$}  & \multicolumn{1}{c}{$333$} & \multicolumn{1}{c}{$134$}    & \multicolumn{1}{c}{$601$}                         \\
    key phrase        & \multicolumn{1}{c}{$38$}   & \multicolumn{1}{c}{$38$} & \multicolumn{1}{c}{$38$}    & \multicolumn{1}{c}{$63 + 80 + 104$}                   \\
    \bottomrule
  \end{tabular}
\end{table}
The resulting set of annotations consists of 114 key phrases, with 38 belonging to each sarcasm category.
Each key phrase within each sarcasm type has at least two non-sarcastic-sounding counterparts. In total, 361 audio samples are manually labeled for analysis.

\begin{table*}[ht!]
  \caption{Mean values of acoustic features}
  \label{tab:acoustic_features}
  \centering
      \resizebox{\textwidth}{!}{
  \begin{tabular}{llllllllll}
    \toprule
    \textbf{} & \textbf{Sarcasm type}  &                
     \multicolumn{2}{c}{\textbf{Mean $F0$}} &
     \multicolumn{2}{c}{\textbf{$F0$ range}} &
     \multicolumn{2}{c}{\textbf{Mean Amplitude}} &
     \multicolumn{2}{c}{\textbf{Duration}}    
     \\
     & &
     \multicolumn{1}{c}{Female} & \multicolumn{1}{c}{Male} &
     \multicolumn{1}{c}{Female} & \multicolumn{1}{c}{Male} &
     \multicolumn{1}{c}{Female} & \multicolumn{1}{c}{Male} &
     \multicolumn{1}{c}{Female} & \multicolumn{1}{c}{Male}    
     \\
    \midrule
    Utterance   & NONE    & \multicolumn{1}{c}{$273.4$ $(60.8)$}  & \multicolumn{1}{c}{$176.9$ $(51.8)$} & \multicolumn{1}{c}{$236.2$ $(88.1)$}    & \multicolumn{1}{c}{$191.0$ $(100.8)$} & \multicolumn{1}{c}{$57.3$ $(7.5)$} & \multicolumn{1}{c}{$60.1$ $(7.4)$} & \multicolumn{1}{c}{$-$} & \multicolumn{1}{c}{$-$}      \\
                & EMB    & \multicolumn{1}{c}{$235.7$ $(46.5)$}  & \multicolumn{1}{c}{$157.1$ $(36.4)$} & \multicolumn{1}{c}{$239.6$ $(83.4)$}    & \multicolumn{1}{c}{$176.5$ $(94.8)$} & \multicolumn{1}{c}{$61.0$ $(6.8)$} & \multicolumn{1}{c} {$63.4$ $(7.4)$} & \multicolumn{1}{c}{$-$} & \multicolumn{1}{c}{$-$}  \\
                & PRO    & \multicolumn{1}{c}{$234.2$ $(50.8)$}  & \multicolumn{1}{c}{$163.9$ $(40.3)$} & \multicolumn{1}{c}{$235.8$ $(98.4)$}    & \multicolumn{1}{c}{$184.9$ $(96.9)$} & \multicolumn{1}{c}{$64.0$ $(9.3)$} & \multicolumn{1}{c}{$64.0$ $(7.0)$} & \multicolumn{1}{c}{$-$} & \multicolumn{1}{c}{$-$}   \\
                & ILL    & \multicolumn{1}{c}{$224.8$ $(56.5)$}  & \multicolumn{1}{c}{$182.5$ $(50.6)$} & \multicolumn{1}{c}{$241.5$ $(102.0)$}    & \multicolumn{1}{c}{$207.3$ $(100.7)$} & \multicolumn{1}{c}{$65.2$ $(11.1)$} & \multicolumn{1}{c}{$61.0$ $(6.7)$} & \multicolumn{1}{c}{$-$} & \multicolumn{1}{c}{$-$}   \\
                \\
    Key phrase   & NONE   & \multicolumn{1}{c}{$210.0$ $(35.7)$}   & \multicolumn{1}{c}{$172.1$ $(58.2)$} & \multicolumn{1}{c}{$79.4$ $(57.8)$}    & \multicolumn{1}{c}{$78.9$ $(84.1)$} & \multicolumn{1}{c}{$57.2$ $(6.3)$}   & \multicolumn{1}{c}{$57.2$ $(5.8)$} & \multicolumn{1}{c}{$0.32$ $(0.13)$} & \multicolumn{1}{c}{$0.40$ $(0.17)$} \\
                & EMB    & \multicolumn{1}{c}{$197.2$ $(27.7)$}  & \multicolumn{1}{c}{$182.8$ $(54.2)$} & \multicolumn{1}{c}{$87.8$ $(77.9)$}    & \multicolumn{1}{c}{$132.3$ $(113.7)$} & \multicolumn{1}{c}{$57.4$ $(8.2)$} & \multicolumn{1}{c}{$62.6$ $(8.3)$} & \multicolumn{1}{c}{$0.35$ $(0.12)$} & \multicolumn{1}{c}{$0.42$ $(0.15)$}   \\
                & PRO    & \multicolumn{1}{c}{$203.4$ $(23.0)$}  & \multicolumn{1}{c}{$177.2$ $(71.1)$} & \multicolumn{1}{c}{$54.7$ $(29.4)$}    & \multicolumn{1}{c}{$98.2$ $(75.9)$} & \multicolumn{1}{c}{$55.8$ $(2.2)$} & \multicolumn{1}{c}{$58.1$ $(5.2)$} & \multicolumn{1}{c}{$0.50$ $(0.20)$} & \multicolumn{1}{c}{$0.60$ $(0.21)$}   \\
                 & ILL    & \multicolumn{1}{c}{$179.8$ $(45.2)$}  & \multicolumn{1}{c}{$143.7$ $(48.9)$} & \multicolumn{1}{c}{$102.4$ $(65.2)$}    & \multicolumn{1}{c}{$108.7$ $(102.3)$} & \multicolumn{1}{c}{$63.8$ $(6.8)$} & \multicolumn{1}{c} {$69.7$ $(6.9)$} & \multicolumn{1}{c}{$0.40$ $(0.16)$} & \multicolumn{1}{c}{$0.45$ $(0.18)$}  \\
    \bottomrule
  \end{tabular}}
\end{table*}

\begin{figure*}[htbp]
  \centering
    \includegraphics[width=\linewidth]{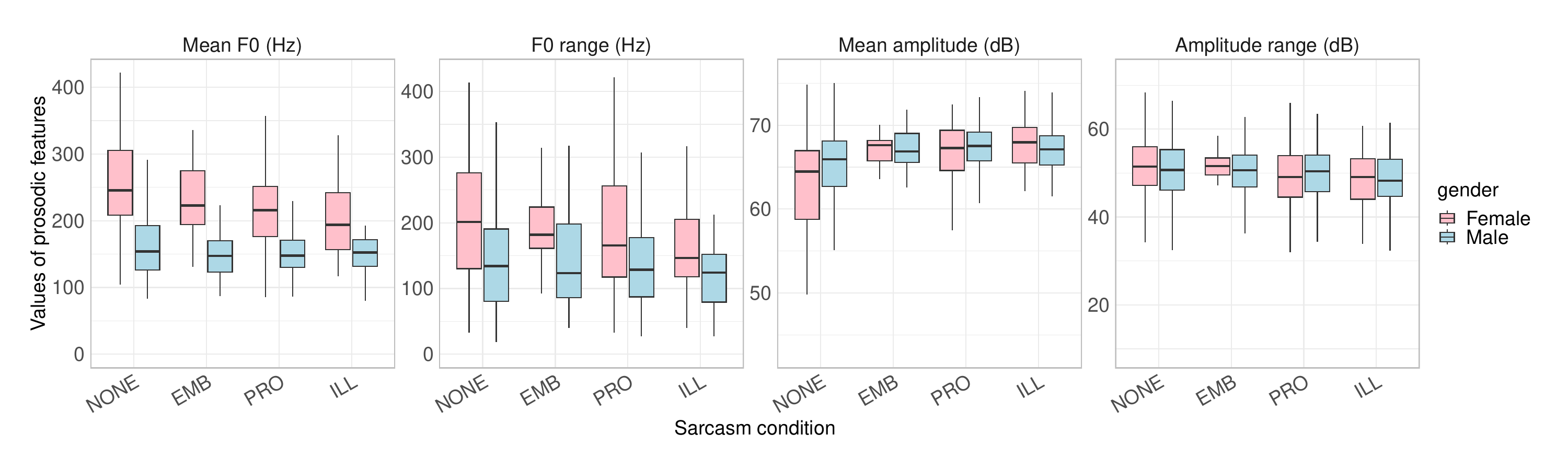}
  \caption{Utterance-level prosodic features across neutral (NONE) and the three sarcastic (EMB, PRO, and ILL) types.}
  \label{fig:uttr_level}
\end{figure*}

\subsection{Acoustic analyses}
In order to find the accurate timestamp for each key phrase, we use a transcription alignment tool Gentle \footnote{https://github.com/lowerquality/gentle} which automatically aligns a transcript with the audio by extracting word-level timestamps from the audio. All validated utterances underwent acoustic analysis with Praat speech analysis software \cite{boersma2011praat}. Building on previous research \cite{cheang2008sound}, we selected several acoustic parameters known for their relevance in distinguishing sarcasm from neutral or sincere expressions. 
The specific measures extracted from each utterance included mean $F0$ (in Hz), $F0$ range (in Hz), mean amplitude (in dB), and amplitude range (in dB). Speech rate is excluded due to the pauses and laughters involved in the audio. However, for each keyword, we extracted duration. 

$F0$ extraction was performed using the auto-correlation method with recommended $F0$ ranges (75-300 Hz for male and 75-500 Hz for female) through Praat. The raw $F0$ points were linearly interpolated and then smoothed. 

\subsection{Statistical analyses}

Based on these measures, we fit linear mixed effects models using \textit{lme4} in R \cite{r, lme} to analyze the effects of the experimental variables on the acoustic measurements. 
Model significance was assessed with the lmerTest package \cite{kuznetsova2017lmertest}. 


For utterance level analysis, we included MESSAGE\_TYPE (PRO: propositional sarcasm, EMB: embedded sarcasm, ILL: illocutionary sarcasm, and NONE: non-sarcasm) and GENDER (male, female) as the fixed factors and SPEAKER as the random factor. For keyword-level analysis, we also included UTTERANCE as the random factor.  

\section{Results and Discussion}
\label{section:results}
The mean measures of $F0$, $F0$ range, amplitude, and duration characterizing different types of utterances and key phrases are summarized in Table~\ref{tab:acoustic_features}. 
Since no significant differences have been found in the amplitude range at both utterance and phrase levels, this feature is omitted in Table 2 for clarity. 

\subsection{Utterance-level features}
As shown in Figure~\ref{fig:uttr_level} and tested by statistical models, both sarcasm type and gender have significant effects on mean $F0$. Specifically, utterances expressing sarcasm have significantly lower mean $F0$ compared to non-sarcastic utterances (EMB: $\beta = -26.94, SE = 10.31, t = -2.61, p < .01$; PRO: $\beta = -24.21, SE = 5.92, t = -4.09, p < .01$; ILL: $\beta = -26.04, SE = 10.31, t = -2.61, p < .01$). As for $F0$ range, no significant differences have been found between sarcastic and non-sarcastic utterance types. In addition, sarcasm type has significant effects on mean amplitude. Utterances expressing sarcasm have significantly higher mean amplitude compared to non-sarcastic utterances (EMB: $\beta = 3.57, SE = 1.50, t = 2.39, p < .05$; PRO: $\beta = 3.85, SE = 0.86, t = 4.48, p < .001$; ILL: $\beta = 3.14, SE = 1.17, t = 2.68, p < .01$).

No systematic differences have been found in $F0$ range and amplitude range across sarcasm types.
At the utterance level, pair-wise comparisons of the three sarcasm types (EMB, PRO, ILL) show no significant difference between the acoustic features. Therefore, we observe no functional trade-off in terms of individual prosodic cues at the utterance level. 

To summarize, sarcasm can be distinguished from neutrality based on prosodic patterns, with sarcastic expressions marked by a lower mean $F0$ and a higher mean amplitude at the utterance level irrespective of the speaker's gender and the context of these expressions, which are in general consistent with previous findings across languages \cite{rockwell2000lower, cheang2008sound, cheang2009acoustic}. 

\begin{figure*}[t]
  \centering
  \includegraphics[width=\linewidth]{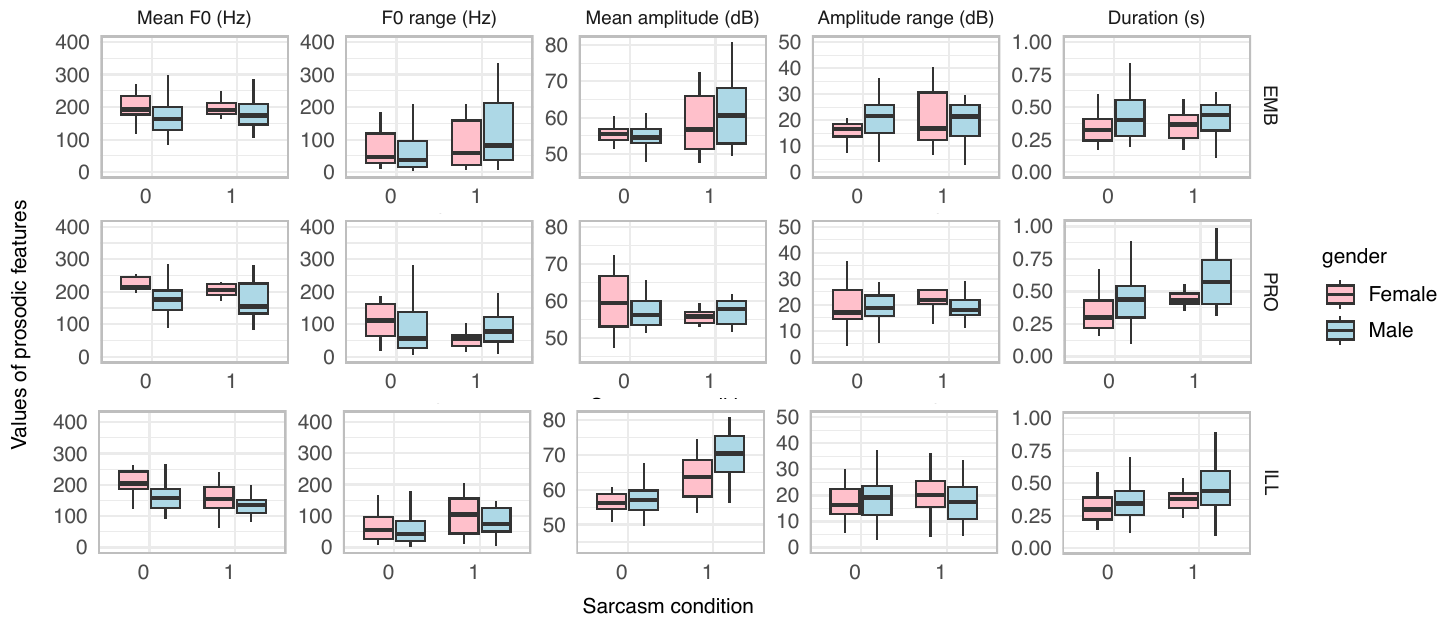}
  \caption{Key phrase-level prosodic features of the three sarcasm types (EMB, PRO, and ILL). ``1" indicates sarcastic key phrases, whereas ``0" signals their non-sarcastic counterparts.}
  \label{fig:word_plot}
\end{figure*}

\subsection{Key phrase-level features}

As shown in Figure~\ref{fig:word_plot} and verified by statistical testing, prosodic characteristics between key phrases extracted from EMB sarcasm utterances and those without sarcasm show no significant differences. 

In comparison, detailed analyses of acoustic features between key phrases extracted from ILL sarcasm utterances and those without sarcasm reveal some notable distinctions. Specifically, ILL sarcasm significantly prolongs the duration of key phrases ($\beta = 0.09, SE = 0.05, t = 1.88, p < .05$), lowers mean $F0$ ($\beta = -27.67, SE = 16.47, t = -1.68, p < .1$), and raises the amplitude level ($\beta = 7.39, SE = 2.09, t = 3.54, p < .001$). However, it does not significantly influence the $F0$ range and the amplitude range. These results are compatible with previous studies on the use of a reduced mean $F0$, a longer duration or a slower speech rate in sarcastic utterances and key words \cite{chen2018s, cheang2008sound}. In addition, we add evidence that mean amplitude may also play an important role in conveying sarcasm, especially in conversational interactions. 

The comparison of prosodic features between key phrases extracted from PRO sarcasm utterances and their neutral counterparts also reveals some differences. Specifically, PRO sarcasm reduces the $F0$ span of key phrases ($\beta = -53.93, SE = 32.11, t = -1.68, p < .1$). However, it does not exert a significant influence on mean $F0$, mean amplitude, amplitude span, or duration. The PRO sarcasm phrases being realized with a relatively reduced $F0$ span than their counterparts echos \cite{chen2018s}'s observation of a flatter pitch fall for sarcastic keywords. Detailed dynamic prosodic measurements of pitch contours could be a promising avenue for future comparisons of three sarcastic types.


\subsection{Trade-off between prosodic and semantic cues}
Researchers agree that prosody is crucial for conveying sarcasm in speech, but views differ in 
how prosody operates independently and interactively with other cues to convey sarcasm \cite{scherer1984vocal, ladd1985evidence}. 
In this study, though no notable differences have been found in the use of prosody in different utterance types, key phrase-level results indicate that speakers tend to rely less on prosodic cues when conveying sarcasm in utterances that are semantically rich in expressing intended sarcasm (i.e., EMB sarcasm type), compared to utterances with fewer semantic or contextual cues (i.e., ILL sarcasm type). This suggests that speakers tend to provide a set of supplementary acoustic cues or enriched prosodic signals in contexts lacking sarcastic semantics to ensure that the speech is unambiguously treated as sarcastic by the listener when semantic features are less indicative of this intent. In addition, different findings observed at the utterance level and the key phrase level illustrate the importance of combining fine-grained and coarse-grained analysis of prosodic features.


Importantly, our study shed light on aspects of the prosodic expression of sarcasm that have not been studied in previous work. The observed functional trade-off between prosodic and semantic cues to sarcasm at the level of key phrases is consistent with the hypothesis postulated and tested in \cite{chen2018s}, though they focused on different utterance types in terms of syntax instead of semantics.

\section{Conclusions} 
\label{section:conclusion}

Using a dataset labeled with different types of sarcasm, we examined the prosodic features for the first time across different sarcastic categories, both at the utterance and the key phrase level. We found significant differences between sarcastic and non-sarcastic utterances in terms of the mean $F0$ and the amplitude level. Furthermore, at the key phrase level, we identified a trade-off between prosodic and semantic cues across different sarcastic categories. 
One limitation of the current study is that annotating non-sarcastic counterparts is time-consuming and error-prone, and only limited data resources are available for each sarcasm type, which imposes constraints on model application. 
Nevertheless, the general findings based on realistic sitcoms instead of lab-recorded data could yield valuable insights into keyword-based controllable speech synthesis and help us further explore the acoustic features that are meaningful and interpretable for multimodal sarcasm detection \cite{gao22f_interspeech, li2023sarcasticspeech}.

\end{document}